\documentclass[pmlr,twocolumn,10pt]{jmlr} 
\usepackage{booktabs}
\usepackage{siunitx}
\usepackage{multirow}
\usepackage[switch]{lineno}
\usepackage{xcolor}

\theorembodyfont{\upshape}
\theoremheaderfont{\scshape}
\theorempostheader{:}
\theoremsep{\newline}

\jmlrvolume{287}
\jmlryear{2025}
\jmlrsubmitted{} 
\jmlrpublished{} 
\jmlrworkshop{Conference on Health, Inference, and Learning (CHIL) 2025}

\title{Electrocardiogram–Language Model for Few-Shot Question Answering with Meta Learning}
\author{%
\Name{Jialu Tang}\Email{j.tang@tue.nl}\\
\addr Eindhoven University of Technology, The Netherlands
\AND
\Name{Tong Xia}\Email{tx229@cam.ac.uk}\\
\addr University of Cambridge, United Kingdom
\AND
\Name{Yuan Lu}\Email{y.lu@tue.nl}\\
\addr  Eindhoven University of Technology, The Netherlands
\AND
\Name{Cecilia Mascolo}\Email{cm542@cam.ac.uk}\\
\addr  University of Cambridge, United Kingdom
\AND
\Name{Aaqib Saeed}\Email{a.saeed@tue.nl}\\
\addr Eindhoven University of Technology, The Netherlands \\
Eindhoven Artificial Intelligence Systems Institute, The Netherlands
}

\begin{document}
\maketitle
\begin{abstract}
Electrocardiogram (ECG) interpretation requires specialized expertise, often involving synthesizing insights from ECG signals with complex clinical queries posed in natural language. The scarcity of labeled ECG data coupled with the diverse nature of clinical inquiries presents a significant challenge for developing robust and adaptable ECG diagnostic systems. This work introduces a novel multimodal meta-learning method for few-shot ECG question answering, addressing the challenge of limited labeled data while leveraging the rich knowledge encoded within large language models (LLMs). Our LLM-agnostic approach integrates a pre-trained ECG encoder with a frozen LLM (e.g., LLaMA and Gemma) via a trainable fusion module, enabling the language model to reason about ECG data and generate clinically meaningful answers. Extensive experiments demonstrate superior generalization to unseen diagnostic tasks compared to supervised baselines, achieving notable performance even with limited ECG leads. For instance, in a 5-way 5-shot setting, our method using LLaMA-3.1-8B achieves an accuracy of 84.6\%, 77.3\%, and 69.6\% on single \textit{verify}, \textit{choose} and \textit{query} question types, respectively. These results highlight the potential of our method to enhance clinical ECG interpretation by combining signal processing with the nuanced language understanding capabilities of LLMs, particularly in data-constrained scenarios.
\end{abstract}

\noindent\textbf{Data and Code Availability} This paper uses ECG-QA, the question-answering dataset for electrocardiogram (ECG) analysis \citep{oh2024ecg} regarding PTB-XL~\citep{wagner2020ptb} and MIMIC-IV-ECG~\citep{gow2023mimic}, separately available at \url{https://github.com/Jwoo5/ecg-qa},  \url{https://physionet.org/content/ptb-xl/1.0.1/}, \url{https://physionet.org/content/mimic-iv-ecg/1.0/}.

\noindent\textbf{Institutional Review Board (IRB)} Our research uses publicly available data, which does not require IRB approval.

\section{Introduction}
\label{sec:introduction}

Electrocardiograms (ECGs) provide a wealth of physiological information crucial for diagnosing a wide range of cardiac conditions. Although doctors are professionally trained to diagnose~\citep{garcia200112,o2008complete}, and even AI systems have shown promise in not only enhancing diagnostic accuracy but also relieving the pressure on healthcare professionals~\citep{jin2024cardiologist,ribeiro2020automatic,hannun2019cardiologist}. However, they are usually trained in limited and incomplete categories~\citep{al2024comprehensive}.
The advent of large language models (LLMs) coupled with advancements in multimodal learning presents a transformative opportunity to enhance ECG interpretation by integrating the rich contextual understanding of language with the detailed physiological insights encoded within ECG signals. This fusion of modalities allows for a more comprehensive and nuanced analysis, potentially leading to more accurate and timely diagnoses. Multimodal question answering (QA) systems, operating at this intersection of ECG data and natural language processing, are emerging as a powerful tool for automating and augmenting clinical workflows, offering the potential to improve diagnostic accuracy, efficiency, and accessibility. By enabling direct interaction with ECG data through natural language queries, these systems can streamline the diagnostic process and empower clinicians with more informed decision-making capabilities.

Developing robust and reliable multimodal QA systems for ECG interpretation relies on the availability of both high-quality and large quantities of labeled data. Yet, obtaining massive amounts of labeled ECGs from cardiologists is costly, which often results in limited datasets. Traditional supervised learning methods tend to perform well only on data with the same distribution as the training data. In real-world deployment, however, models frequently encounter new tasks and previously unseen populations outside the training distribution, where traditional methods may fail. Meta-learning~\citep{andrychowicz2016learning, finn2017model, thrun1998learning}, a paradigm focused on ``learning to learn'', offers a compelling solution to this challenge. By training models on a diverse range of tasks, meta-learning enables them to acquire transferable knowledge and adapt rapidly to new, unseen tasks with minimal labeled data. This adaptive capacity is particularly valuable in the ECG-language QA domain, where new diagnostic questions and data distributions constantly emerge.

\begin{table*}[t]
\centering
\caption{Overview of question types and data distribution within the meta learning benchmark dataset created for few-shot ECG question answering.}

\resizebox{\textwidth}{!}{%
\begin{tabular}{@{}lccccl@{}}
\toprule
\textbf{Question Type} &
  \textbf{Attributes} &
  \textbf{Answers} &
  \textbf{Classes (train:test)} &
  \textbf{Samples} &
  \multicolumn{1}{c}{\textbf{Example}} \\ \midrule
Single-Verify &
  94 &
  yes/no &
  156 (124:32) & 34,105
  &
  \begin{tabular}[c]{@{}l@{}}Q: Does this ECG show 1st degree av block? \\ A: yes/no\end{tabular} \\
Single-Choose &
  165 &
  both/none/attr\_1/attr\_2 &
  262 (209:53) & 47,655
  &
  \begin{tabular}[c]{@{}l@{}}Q:  Which noise does this ECG show, baseline drift or static noise?\\ A: baseline drift /static noise/both/none\end{tabular} \\
Single-Query &
  30 &
  attr\_1/attr\_2/\ldots/attr\_n &
  260 (208:52) & 63,125
  &
  \begin{tabular}[c]{@{}l@{}}Q: What direction is this ECG deviated to?\\ A: Normal axis/ \ldots./open-ending\end{tabular} \\
All &
  206 &
  yes/no/both/none/\ldots/attr\_n &
  678 (541:137) & 144,885
  &
  \multicolumn{1}{c}{\ldots} \\ \bottomrule
\end{tabular}%
}
\label{tab:data_statistics}
\end{table*}

Few-shot learning (FSL), as a practical approach within meta-learning, shows significant promise in various medical imaging tasks \citep{pachetti2024systematic}. The success of FSL underscores the potential of learning efficient representations that generalize effectively from limited examples \citep{finn2017model}. While high-quality multimodal datasets, like those available for chest X-rays, have fueled progress in FSL for image-based diagnostics, the ECG domain lacks datasets specifically tailored for few-shot multimodal learning paradigms. The recent introduction of the ECG-QA dataset~\citep{oh2024ecg}, built upon established ECG resources like PTB-XL~\citep{wagner2020ptb} and MIMIC-IV-ECG~\citep{gow2023mimic}, partially addresses this need with its diverse question types (single-verify, single-choose, single-query) and ECG attributes (e.g., SCP codes, noise types, heart axis deviations). However, existing dataset lacks the structured task configurations necessary for developing and evaluating meta-learning models, leaving a significant gap in the advancement of ECG-language QA systems.

In response to these challenges, we propose a novel, LLM-agnostic, multimodal meta-learning framework specifically designed for few-shot ECG-language QA. Our architecture integrates a self-supervised pre-trained ECG encoder with a frozen LLM and a trainable multimodal fusion mapper bridging the ECG and language representations. This fusion mapper is crucial for acquiring transferable meta-knowledge, enabling rapid adaptation to new tasks. Furthermore, we create a benchmarking variant of the ECG-QA dataset (see Table~\ref{tab:data_statistics}), designed to facilitate meta-learning having diverse tasks with varying attribute-answer combinations. This benchmark dataset allows us to rigorously evaluate a model's ability to generalize to unseen diagnostic tasks in a few-shot setting. We demonstrate the effectiveness of our framework across a broad range of language models, showcasing superior generalization performance compared to fully supervised baselines in various few-shot settings and question types. Our findings highlight the potential of our approach to significantly impact clinical practice by enabling robust and adaptable ECG-language QA with limited labeled data.

\section{Related Works}
\begin{figure*}[t]
\centering
\includegraphics[width=1\textwidth]{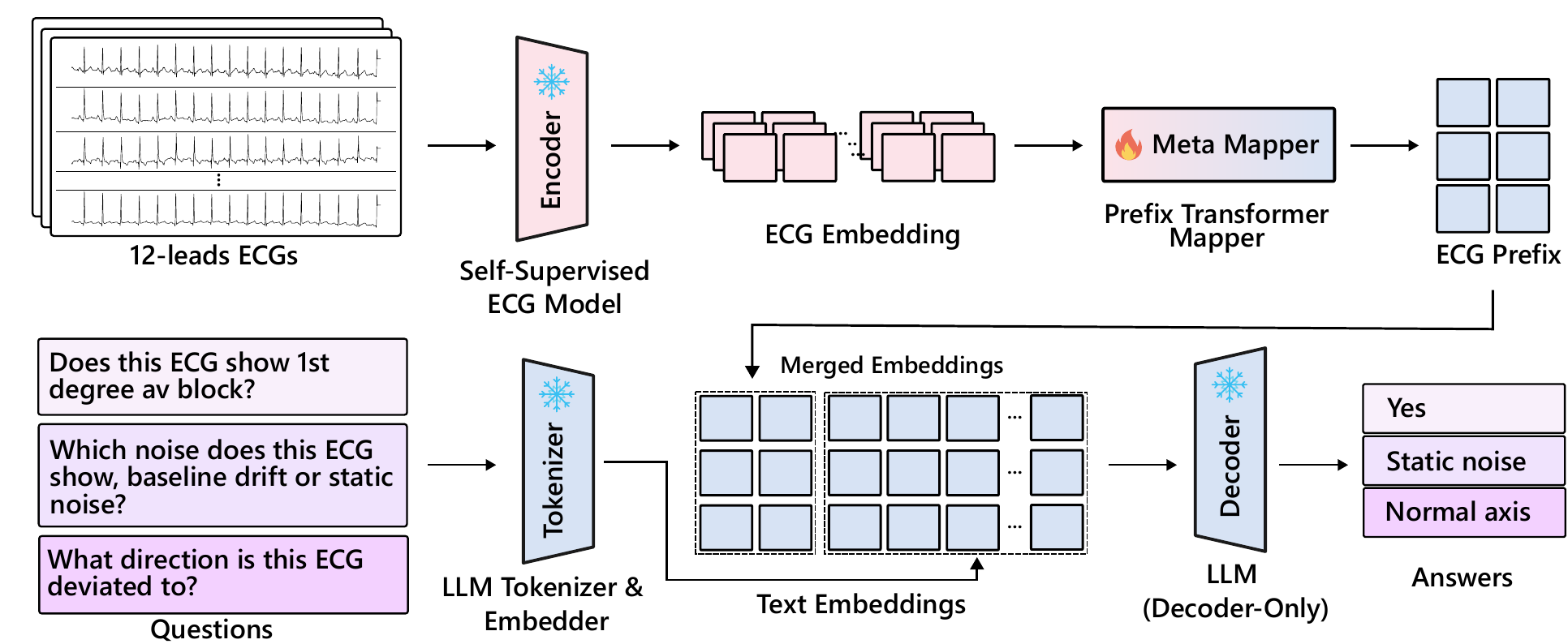}
\caption{Overview of our proposed multimodal few-shot ECG question answering approach, integrating ECG signals and textual queries via a fusion module for a frozen LLM to generate answer in a natural language.}
\label{fig:model}
\end{figure*}

Deep learning has significantly advanced ECG interpretation, with models such as CNNs and Transformers demonstrating promising results in automated diagnosis \citep{chugh2023systematic, woo2024unified, sun2023artificial}. However, these supervised approaches typically require large labeled datasets, hindering their generalizability to diverse patient populations and uncommon ECG presentations, a critical limitation in real-world clinical settings. While self-supervised learning methods offer a potential solution by learning from unlabeled ECG data \citep{gopal20213kg, tonekaboni2021unsupervised, oh2022lead, saeed2019multi,kiyasseh2021clocs}, they have not yet been effectively leveraged for complex clinical question answering involving nuanced language understanding.

Multimodal learning has emerged as a powerful paradigm in healthcare, demonstrating success in integrating medical images with textual information \citep{KRONES2025102690, boecking2022making, zhang2020combining, rasmy2021med, warner2024multimodal, acosta2022multimodal}. However, effectively fusing temporal physiological signals like ECG with the unstructured and often ambiguous nature of clinical language presents unique challenges, particularly in generative tasks like open-ended question answering. Our work directly addresses this gap by proposing a novel method for ECG-language fusion, enabling more comprehensive and nuanced diagnostic reasoning by leveraging the complementary information present in both modalities.

Furthermore, the inherent scarcity of labeled data for specific cardiac conditions necessitates efficient few-shot learning strategies. Meta-learning techniques, such as MAML \citep{finn2017model}, have shown promise in enabling rapid adaptation to new tasks with limited examples \citep{vettoruzzo2024advances}, offering a compelling approach for ECG interpretation. While recent studies have explored integrating LLMs with few-shot learning in medical domains \citep{jin2023time, yu2023zero}, the potential of combining meta-learning, LLMs, and multimodal fusion for ECG-language question answering remains largely unexplored. Our work contributes a method that integrates these key components, enabling adaptability to new tasks from limited labeled data while leveraging the powerful language understanding and generation capabilities of LLMs.

\section{Methods}
We present a method capable of rapidly adapting models to novel ECG Question-Answers (QAs) tasks with minimal labeled data. We frame this problem within the context of multimodal few-shot meta-learning consisting of three key phases. Here, we first define the meta-learning dataset specific to ECG-language QAs , where the objective is to classify unseen examples into one of N new ‘test’ classes, given only a few reference examples per class\citep{triantafillou2019meta}. Then, we detail the architecture of our proposed model that integrates ECG analysis with question processing to generate the corresponding answer, and finally, we describe the procedures for few-shot meta-training and inference,in which a few gradient steps may provide strong results on a new task can be considered as constructing an internal representation that is generically applicable to numerous tasks.\citep{YUAN202353}

\subsection{Problem Formulation}
We focus on the task of ECG-based question answering, where the goal is to predict an answer $a$ given an ECG signal $x$ and a natural language question $q$. Formally, we aim to learn a function $f_\theta$ parameterized by $\theta$: $a = f_\theta(x, q)$. Due to the scarcity of labeled data for certain ECG conditions and the need to generalize to emerging diseases, we adopt a few-shot learning approach. In this setting, we have access to a set of tasks, each consisting of a small support set and a query set. A single meta-learning step refers to an optimization after a support set (i.e., the few-shot samples) is used by the model to learn across different tasks and a query set adapts to a new task \citep{ravi2016optimization}. Specifically, let $\mathcal{D}_{\text{meta-train}}$ denote the meta-training dataset comprising $n$ tasks $\{\mathcal{T}_1, \mathcal{T}_2, \dots, \mathcal{T}_n\}$, where each task $\mathcal{T}_i$ consists of a support set $D^\text{s}_i$ and a query set $D^\text{q}_i$: $\mathcal{T}_i = (D^\text{s}_i, D^\text{q}_i)$.

In the $N$-way $K$-shot setting, the support set $D^\text{s}_i$ contains $N$ classes (attribute-answer pairs), each with $K$ labeled examples. Each example in the support and query sets is a triplet $(x, q, a)$, where $x$ is an ECG signal, $q$ is a question about $x$, and $a$ is the corresponding answer. Our objective is to train a model that can, given the support set $D^\text{s}_i$ of a new task $\mathcal{T}_i$, adapt to accurately predict the answers in the query set $D^\text{q}_i$. This requires the model to generalize to new attribute-answer combinations and diverse question formulations with minimal labeled data.

\paragraph{Meta Learning Benchmark Dataset.}
We create a benchmarking dataset for meta learning in our study using the ECG-QA dataset~\citep{oh2024ecg}, which contains question-answer pairs annotated by expert clinicians and is built upon the PTB-XL \citep{wagner2020ptb} and MIMIC-IV-ECG \citep{gow2023mimic} datasets. We focus on questions involving a single ECG and consider three types of questions:
\begin{itemize}
    \item \textbf{Single-Verify}: Yes/no questions, e.g., ``Does this ECG show atrial fibrillation?''
    \item \textbf{Single-Choose}: Multiple-choice questions selecting from two or more options, e.g., ``Which type of noise is present in this ECG: baseline drift or muscle artifact?''
    \item \textbf{Single-Query}: Open-ended questions requiring specific attribute values, e.g., ``What is the heart axis direction in this ECG?''
\end{itemize}

We create our dataset for few-shot meta learning by categorizing questions based on six types of attributes: \textit{SCP codes}, \textit{noise types}, \textit{stages of infarction}, \textit{presence of ectopic beats}, \textit{heart axis deviations}, and \textit{numeric features}. Each attribute encompasses multiple sub-attributes, leading to a diverse set of attribute-answer pairs. For instance, the \textit{SCP codes} attribute includes specific diagnoses such as ``non-diagnostic T-wave abnormalities`` and ``conduction disturbances``.

Each class in our few-shot learning tasks corresponds to a unique attribute-answer pair. For the \textit{Single-Verify} questions, classes are formed by pairs of attributes and binary answers (\textit{yes/no}). Similarly, for \textit{Single-Choose} questions, classes are based on attributes and possible options (\textit{both}, \textit{none}, specific sub-attributes), and for \textit{Single-Query} questions, classes are defined by attributes and their specific values.

We construct the meta-training dataset $\mathcal{D}_{\text{meta-train}}$ and the meta-testing dataset $\mathcal{D}_{\text{meta-test}}$ with mutually exclusive classes to evaluate the model's ability to generalize to unseen attribute-answer pairs. Table~\ref{tab:data_statistics} summarizes the number of attributes, possible answers, and classes for training and testing datasets in each question type. 
To ensure diversity and robustness, we include multiple question formulations with the same meaning but diverse expressions within each class. For example, the questions "Is non-diagnostic T-wave abnormality present in this ECG?" and "Does this ECG reveal signs of non-diagnostic T-wave abnormalities?" belong to the same class but provide variability in the language.

\paragraph{Task Definition.}
Formally, let $\mathcal{D}_{\text{meta-train}}$ be the set of meta-training data, defined as: $\mathcal{D}_{\text{meta-train}} = \{(D_1^s, D_1^q), (D_2^s, D_2^q), \ldots, (D_n^s, D_n^q)\}$. In the context of few-shot learning, $N$-way refers to the number of distinct attribute-answer pair classes in each task. The support set $D_i^s$ for the $i$-th task is defined as: $D_i^s = \bigcup_{c=1}^N D_{i,c}$
where $D_{i,c}$ represents the set of $K$ labeled examples for the $c$-th class in the support set: $D_{i,c} = \{S_{i,c}^{(1)}, S_{i,c}^{(2)}, \ldots, S_{i,c}^{(K)}\}$
Each sample $S_{i,c}^{(j)}$ is defined as: $S_{i,c}^{(j)} = (x_{i,c}^{(j)}, q_{i,c}^{(j)}, a_{i,c}^{(j)})$
\noindent where $x_{i,c}^{(j)}$ is the ECG signal, $q_{i,c}^{(j)}$ is the input question text, and $a_{i,c}^{(j)}$ is the corresponding answer text. The query set $D_i^q$ contains additional samples from the same classes, with $M$ query samples per class ($M > K$), where $K$ represents the number of ways in few-shot learning setting. This formulation tests the model's ability to generalize to unseen ECGs and diverse question expressions within the same attribute-answer classes.

\subsection{Model Architecture}
\label{sec:ecg_encoder}
\label{sec:model_architecture}
 The architecture for ECG-based question answering consists of four main components: 
(1) a pretrained and frozen text tokenizer and embedder for semantic understanding of questions, (2) a pretrained and frozen ECG encoder for extracting meaningful representations from ECG signals, (3) a trainable multimodal fusion module to align ECG embeddings with the textual representation space, and (4) a text decoder to generate language-based answers, as illustrated in Figure~\ref{fig:model}.

\paragraph{Text Encoder.}
\label{sec:text_encoder}
We employ a Transformer-based large language model to tokenize and embed the input textual data. Given a set of questions 
$Q = \{q_1, q_2, \ldots, q_N\}$ and corresponding answers 
$A = \{a_1, a_2, \ldots, a_N\}$, each question $q_i$ is tokenized into a sequence of embeddings $S_i = \{s_{i,1}, s_{i,2}, \ldots, s_{i,L}\}$, where $L$ denotes the length of the tokenized question. 

\paragraph{ECG Encoder.}
To extract meaningful representations from ECG signals, we pre-train an ECG encoder based on prior work~\citep{oh2022lead}. Let $X = \{x_1, x_2, \ldots, x_N\}$ represent a set of ECG recordings, where each $x_i \in \mathbb{R}^{T_s \times C}$ corresponds to an ECG signal with $T_s$ time steps and $C$ leads. The ECG encoder processes each $x_i$ to produce a sequence of embeddings $E_i = \{e_{i,1}, e_{i,2}, \ldots, e_{i,K}\}$, capturing both local and global features of the ECG data.

The encoder incorporates techniques such as Wav2Vec (W2V), Contrastive Masked Segment Comparison (CMSC), and Random Lead Masking (RLM)\citep{oh2022lead}, pre-trained on the PhysioNet 2021 dataset \citep{PhysioNet}. The W2V component uses convolutional and Transformer layers to derive contextualized representations from raw ECG signals. CMSC enhances temporal invariance by contrasting adjacent segments within ECG recordings. RLM improves generalization by masking random leads during training, enabling robustness across varying lead configurations.

\paragraph{Multimodal Fusion Mapper (Meta Mapper).}

The multimodal fusion module integrates textual and ECG representations to generate a joint embedding for question answering. We transform the ECG embeddings $E_i$ into a prefix embedding $P_i = \{p_{i,1}, p_{i,2}, \ldots, p_{i,M}\}$ that aligns with the dimensionality of the question embeddings $S_i$. This is achieved through a transformation network that projects $E_i$ into the same embedding space as $S_i$.

Specifically, we apply linear transformations to $E_i$ to obtain query ($Q$), key ($K$), and value ($V$) matrices, enabling an attention mechanism defined as: $\text{Attention}(Q, K, V) = \text{softmax}\left(\frac{QK^\top}{\sqrt{d_k}}\right)V$, where $d_k$ is the dimensionality of the key vectors. This attention mechanism captures interactions between ECG features and the textual context, facilitating effective multimodal fusion. The fusion module's parameters are trainable during meta-learning, allowing adaptation to new tasks.

\paragraph{Text Decoder (Language Model).}
\label{sec:text_decoder}
The text decoder generates the answer $a_i$ based on the concatenated embeddings of the ECG prefix $P_i$ and the tokenized question $S_i$. Using a Transformer-based language model, the decoder autoregressively produces the answer tokens until an end-of-sequence token is reached or a maximum length is exceeded. By integrating the ECG encoder with the language model through the multimodal fusion module, our architecture effectively leverages both physiological signals and textual information to address the multimodal question-answering task in a few-shot learning setting.

\subsection{Few-shot Meta Training and Inference.}
To enable rapid adaptation to new ECG question-answering tasks with minimal labeled data, we employ a few-shot meta-learning technique based on Model-Agnostic Meta-Learning (MAML)\citep{finn2017model}. The meta-training process aims to find model parameters that are well-suited for quick fine-tuning on unseen tasks.

\paragraph{Meta-Training Phase.}
During meta-training as shown in Appendix \ref{apd:second} Figure \ref{fig:meta_training_overview}, we sample a batch of tasks ${\mathcal{T}_i}$ from the task distribution $p(\mathcal{T})$. Each task ${\mathcal{T}_i}$ consists of a support set $D_i^s$ and a query set $D_i^q$. The support set contains $N$ classes with $K$ examples each ($N$-way $K$-shot learning), and the query set is used to evaluate adaptation performance.

\paragraph{Inner Loop: Task Adaptation}
For each task ${\mathcal{T}_i}$, we perform adaptation by minimizing the task-specific loss $L_{\mathcal{T}_i}$ on the support set $D_i^s$:
\[
\theta_i' = \theta - \alpha \nabla_\theta L_{\mathcal{T}_i}(f_\theta; D_i^s)
\]

\noindent where $\theta$ are the model parameters, $\theta_i'$ are the adapted parameters for task ${\mathcal{T}_i}$, $\alpha$ is the inner-loop learning rate, and $f_\theta$ denotes the model. The loss $L_{\mathcal{T}_i}$ is computed using the negative log-likelihood over the support set:
\[
L_{\mathcal{T}_i}(f_\theta; D_i^s) = - \sum_{(x_j, q_j, a_j) \in D_i^s} \log p(a_j | x_j, q_j; \theta)
\]

\paragraph{Outer Loop: Meta-Optimization.}
After adapting to each task, we evaluate the adapted model $f_{\theta_i'}$ on the corresponding query set $D_i^q$ and compute the meta-loss:
\[
L_{\text{meta}}(\theta) = \sum_{\mathcal{T}_i \sim p(\mathcal{T})} L_{\mathcal{T}_i}(f_{\theta_i'}; D_i^q)
\]
The model parameters $\theta$ are updated to minimize the meta-loss using gradient descent:
\[
\theta \leftarrow \theta - \beta \nabla_\theta L_{\text{meta}}(\theta)
\]

\noindent where $\beta$ is the outer-loop learning rate. This update encourages the learned parameters $\theta$ to be easily adaptable to new tasks.

\paragraph{Meta-Testing Phase.}
In the meta-testing phase, we assess the model's ability to adapt to unseen tasks from the meta-test set $D_{\text{meta-test}}$. For each new task $\mathcal{T}_{\text{new}}$, we perform adaptation using the support set $D_{\text{new}}^s$:
\[
\theta_{\text{new}}' = \theta - \alpha \nabla_\theta L_{\mathcal{T}_{\text{new}}}(f_\theta; D_{\text{new}}^s)
\]

The adapted parameters $\theta_{\text{new}}'$ are then utilized to predict answers on the query set $D_{\text{new}}^q$, evaluating the model's generalization to new tasks.

\section{Experiments}

\subsection{Implementation Details}
We utilize a self-supervised pre-training strategy of \citep{oh2022lead} (see Section \ref{sec:ecg_encoder}) for pre-training ECG encoder  using the publicly available PhysioNet 2021 Challenge datasets \citep{PhysioNet}. Each ECG recording is sampled at 500 Hz and has a duration ranging from 5 to 144 seconds. For the global contrastive learning task, we segment each recording into 5-second segments (corresponding to 2,500 samples). The rest of the implementation details are provided in Appendix \ref{apd:Pretraining}.

\subsection{Pre-Processing} 
Due to class imbalance, we exclude data points of classes with fewer than 140 samples for Single-Verify questions, 14 samples for Single-Choose, and 50 for Single-Query question types as described further in Appendix \ref{apd:Dataset}.

\begin{table*}[t]
\centering
\caption{Performance comparison (Accuracy \%) of few-shot and fully-supervised models on multimodal question answering across various question types and few-shot settings (N-way K-shot).}
\label{tab:main_table}
\setlength\tabcolsep{1.8pt}
\resizebox{\textwidth}{!}{%
\begin{tabular}{@{}lccccccc@{}}
\toprule
\multirow{2}{*}{\textbf{Method}} & \multirow{2}{*}{\textbf{Language Model}} & \multirow{2}{*}{\textbf{Episodic}} & \textbf{Few-shot Setting} & \multicolumn{4}{c}{\textbf{Questions Type}} \\ \cmidrule(l){4-8} 
 &  &  & \textbf{N-Way K-Shot} & \textbf{S-Verify} & \textbf{S-Choose} & \textbf{S-Query} & \textbf{All-Single (Combined)} \\ \midrule
\multirow{2}{*}{Baseline (Supervised)} & Gemma-2-2B & $\times$ & N/A & 45.1   & 12.6   & 7.1   & 6.9   \\ 
& Llama-3.1-8B & $\times$ & N/A & 83.8   & 34.8   & 25.4   & 25.0   \\ \midrule
\multirow{4}{*}{Ours} & \multirow{4}{*}{Gemma-2-2B} & \multirow{4}{*}{$\checkmark$} & 2-5 & 89.0   & 84.5   & 48.6   & 41.2   \\
 &  &  & 2-10 & 90.9   & 86.1   & 49.3   & 42.7   \\
 &  &  & 5-5 & 82.4   & 62.9   & 42.1   & 46.2   \\
 &  &  & 5-10 & 83.4   & 65.1   & 52.2   & 48.5   \\ \midrule
\multirow{4}{*}{Ours} & \multirow{4}{*}{Llama-3.1-8B} & \multirow{4}{*}{$\checkmark$} & 2-5 & 90.3   & 81.3   & 63.9   & 62.5   \\
 &  &  & 2-10 & 92.7   & 87.2   & 67.6   & 64.7   \\
 &  &  & 5-5 & 84.6   & 77.3   & 69.6   & 71.1   \\
 &  &  & 5-10 & 85.8   & 79.6   & 73.9   & 75.3   \\ \midrule
~\cite{oh2024ecg} & - & $\times$ & N/A & 74.6   & 57.1   & 41.0   & - \\ 
\bottomrule
\end{tabular}%
}
\end{table*}

\subsection{Multimodal Fusion Module Architecture}
\label{sec:multimodal_fusion_mapper}

We experiment with multiple mapping approaches tailored to different aspects of feature transformation and use Attention-based Mapper (see in Appendix \ref{apd:multimodal_fusion_mapper} ) as a default mechanism due to its superior performance.

\subsection{Training \& Inference Procedures}
The optimization of the meta-learning model is performed using the AdamW optimizer with 10,000 meta-training steps and 1,000 meta-testing steps. Rest of the training details are provided in Appendix \ref{apd:Training infer}.

\subsection{Performance Evaluation}
\label{sec:evaluation}
We assess the model's performance by comparing the overlap between the generated answers and the ground truth. Given that the generated sequences may vary in length from the ground truth, we compute the accuracy by aligning the generated sequence to the length of the ground truth: $\text{Accuracy} = \frac{1}{n} \sum_{i=1}^n I(\hat{a}_i = a_i)$, where, $\hat{a}_i$ is the generated token at position $i$, $a_i$ is the ground truth token at the same position, $n$ is the length of the ground truth sequence, and $\mathbb{I}(\cdot)$ is the indicator function, which equals $1$ if the condition is true and $0$ otherwise. 
Furthermore, we also evaluate the model's performance using various natural language generation (NLG) metrics, including BLEU \citep{papineni2002bleu}, BertScore \citep{zhang2019bertscore}, and Rouge \citep{lin2004rouge} as these have been broadly utilized to evaluate the LLM generated text~\citep{abbasian2024foundation}.

\section{Results}

\begin{table*}[t]
\centering
\caption{Performance comparison (\%) with natural language generation metrics (i.e., BLEU-1, BertScore, and Rouge) of few-shot and supervised (standard) models across question types and few-shot settings (N-way K-shot).}
\setlength\tabcolsep{1.8pt}
\resizebox{\textwidth}{!}{%
\begin{tabular}{lccclcccccccccccc}
\toprule
\multirow{2}{*}{\textbf{Method}} & \multirow{2}{*}{\textbf{Language Model}} & \multirow{2}{*}{\textbf{Episodic}} & \multirow{2}{*}{\textbf{Few-shot Setting}} & \multicolumn{3}{c}{\textbf{Single-Verify}} & \multicolumn{3}{c}{\textbf{Single-Choose}} & \multicolumn{3}{c}{\textbf{Single-Query}} & \multicolumn{3}{c}{\textbf{All-Single}} \\ \cmidrule(lr){5-7} \cmidrule(lr){8-10} \cmidrule(lr){11-13} \cmidrule(lr){14-16}
 &  &  &  & \textbf{BLEU} & \textbf{BertScore} & \textbf{Rouge} & \textbf{BLEU} & \textbf{BertScore} & \textbf{Rouge} & \textbf{BLEU} & \textbf{BertScore} & \textbf{Rouge} & \textbf{BLEU} & \textbf{BertScore} & \textbf{Rouge} \\ \midrule

\multirow{2}{*}{Baseline} & Gemma-2-2B & $\times$ & N/A & 34.4 & 42.8 & 33.9 & 12.4 & 35.8 & 13.0 & 3.2 & 36.7 & 7.5 & 4.9 & 37.2 & 6.9 \\
 & Llama-3.1-8B & $\times$ & N/A & 69.8 & 92.9 & 69.8 & 37.3 & 68.3 & 38.4 & 15.7 & 53.2 & 17.7 & 12.9 & 54.3 & 17.0 \\ \midrule

\multirow{4}{*}{Ours} & \multirow{4}{*}{Gemma-2-2B} & \multirow{4}{*}{$\checkmark$} & 2-5 & 75.8 & 94.3 & 75.8 & 73.4 & 87.4 & 76.4 & 36.0 & 67.2 & 32.8 & 34.9 & 69.5 & 38.9 \\
 &  &  & 2-10 & 78.3 & 94.8 & 78.3 & 73.5 & 87.4 & 75.6 & 38.3 & 70.0 & 46.5 & 35.4 & 71.0 & 39.9 \\
 &  &  & 5-5 & 60.8 & 90.0 & 60.8 & 48.5 & 72.7 & 50.6 & 25.3 & 61.7 & 32.7 & 32.7 & 69.2 & 35.8 \\
 &  &  & 5-10 & 68.2 & 92.1 & 68.2 & 52.6 & 75.4 & 54.2 & 30.1 & 64.8 & 37.5 & 35.0 & 69.7 & 39.6 \\ \midrule

\multirow{4}{*}{Ours} & \multirow{4}{*}{Llama-3.1-8B} & \multirow{4}{*}{$\checkmark$} & 2-5 & 79.9 & 95.2 & 79.9 & 77.8 & 88.8 & 79.3 & 36.3 & 67.6 & 43.7 & 37.8 & 73.1 & 40.9 \\
 &  &  & 2-10 & 81.2 & 95.6 & 81.2 & 77.9 & 89.3 & 79.4 & 43.0 & 71.9 & 49.7 & 42.1 & 73.8 & 46.5 \\
 &  &  & 5-5 & 66.2 & 92.0 & 66.2 & 69.4 & 84.8 & 71.0 & 27.9 & 63.3 & 34.2 & 30.4 & 68.4 & 33.0 \\
 &  &  & 5-10 & 72.8 & 72.8 & 72.8 & 79.6 & 90.2 & 80.7 & 31.0 & 65.4 & 37.7 & 35.2 & 70.2 & 38.5 \\ \bottomrule

\end{tabular}%
}
\label{tab:nlg_metrics_table}
\end{table*}

Here, we evaluate the performance of our approach, analyzing the impact of different design choices and training strategies. We investigate the effectiveness of episodic training, which enables models to quickly adapt to new tasks by simulating distinct tasks  for rapid inner loop learning, compare our few-shot generative approach with a fully supervised classification baseline, assess the influence of model size, analyze the performance of different multimodal fusion mappers, and examine the effects of freezing the ECG encoder parameters. We compare our few-shot generative approach with a fully supervised classification baseline. This comparison assesses the influence of model size, analyzes the performance of different multimodal fusion mappers, and examines the effects of freezing the ECG encoder parameters. Finally, we explore the role of meta-knowledge and evaluate performance across various ECG attributes.

\subsection{Episodic Training and Comparison with Supervised Baselines}
We evaluate the effectiveness of episodic training for few-shot multimodal question answering. Table~\ref{tab:main_table} presents the performance of two large language models (LLMs), Gemma-2-2B \citep{team2024gemma} and Llama-3.1-8B \citep{dubey2024llama}, under various few-shot settings (2-way 5-shot, 2-way 10-shot, 5-way 5-shot, and 5-way 10-shot) and question types (see Table~\ref{tab:data_statistics}, Single-Verify, Single-Choose, Single-Query, and All Single question types). We compare episodic training with standard supervised learning (Baseline) for each LLM. The results demonstrate that episodic training consistently improves performance across all settings and question types, highlighting its ability to generalize to unseen queries. Furthermore, we compare our few-shot generative approach with a fully supervised classification model adapted from image captioning to ECG question answering \citep{oh2024ecg} (Upper Bound), which serves as an upper-bound on the performance. This model was trained on the original ECG-QA dataset \citep{oh2024ecg} and uses exact match accuracy. In contrast, our model's accuracy is measured by the overlap between the ground truth and the generated answer (Section~\ref{sec:evaluation}) and NLG metrics in Table~\ref{tab:nlg_metrics_table} for our key models. Our results also showcase the performance improvement achieved by using a larger LLM (Llama-3.1-8B) compared to a smaller one (Gemma-2-2B).

Furthermore, Appendix \ref{apd:third} Figure~\ref{fig:ecg_qa_examples} provides a comparative analysis of Gemma-2-2B and Llama-3.1-8B on ECG-related question answering tasks. It shows example ECGs (leads II, V1, and V6) alongside representative questions from each of the three question types. For each query, we present the ground truth (GT) and the models' responses (A), enabling a direct visual comparison of their performance. This visualization complements the quantitative results in Table~\ref{tab:main_table}, offering insights into the models' reasoning processes and their ability to extract and articulate information from ECG data across varied question formats.

\subsection{Impact of Model Scale}
We evaluate the 5-way 5-shot setting in single-choose question few-shot performance of several large language models (LLMs) on ECG-language question answering by simply replacing the corresponding LLM in our method, including Gemma-2-2B \citep{team2024gemma}, Llama-3.1-8B \citep{dubey2024llama}, GPT-2 \citep{radford2019language}, Phi-2-2B \citep{javaheripi2023phi}, Qwen-2-1.5B \citep{bai2023qwen}, SmolLM-2-1.7B \citep{allal2025smollm2}, DeepSeek-R1-1.5B \citep{guo2025deepseek}. As shown in Table~\ref{tab:llms}, Llama-3.1-8B consistently achieves the highest accuracy across all question types, demonstrating a substantial performance improvement. Specifically, Llama-3.1-8B exhibits a 2.2\%, 14.4\%, and 27.5\% improvement over the best-performing 2B parameter model (Gemma-2-2B) on S-Verify, S-Choose, and S-Query, respectively, culminating in a 24.9\% overall improvement (All-S). This marked improvement suggests that the increased parameter count of Llama-3.1-8B facilitates the learning of richer representations that better capture nuanced relationships between ECG data and corresponding natural language queries. We hypothesize that utilizing an even larger LLM could potentially lead to further significant performance improvements.

While Llama-3.1-8B exhibits superior performance, its computational requirements are substantial. Within the set of 2B parameter models, Gemma-2-2B demonstrates the strongest performance, offering a compelling balance between accuracy and computational efficiency. Accordingly, we adopt Gemma-2-2B as the default model for subsequent ablation studies.

\begin{table}[t]
\centering
\caption{Comparison (Accuracy \%) of various language models.}
\label{tab:llms}
\tiny
\begin{tabular}{@{}lcccc@{}}
\toprule
\textbf{Language Model} & \textbf{S-Verify} & \textbf{S-Choose} & \textbf{S-Query} & \textbf{All-S} \\ \midrule
GPT-2-1.5B & 72.8   & 47.2   & 19.8   & 23.2   \\
Phi-2-2B & 65.7   & 33.0   & 51.5   & 22.2   \\
SmolLM-2-1.7B & 69.9   & 43.5   &  17.4  & 27.8   \\
DeepSeek-R1-1.5B &  79.9  & 52.7   &  22.1  &  16.1  \\ 
Qwen-2-1.5B & 70.8   & 46.7   & 17.9   & 20.1   \\
Gemma-2-2B & 82.4   & 62.9   & 42.1   & 46.2   \\
Llama-3.1-8B & 84.6   & 77.3   & 69.6   & 71.1   \\ 
\bottomrule
\end{tabular}%
\end{table}

\subsection{Performance Analysis Across Attribute Types}
Table \ref{tab:attribute_type} presents the model's performance across various ECG attribute types for three question types in a 2-way 5-shot setting. Overall, the model demonstrates strong performance across the different attribute types. The model achieves particularly high accuracy for the SCP Code attribute across the board, potentially attributable to the larger amount of training data available for this type. Conversely, performance on attributes like extra systole exhibits greater variability, particularly in the single-choose task, suggesting inherent challenges associated with this attribute. The observed differences in accuracy across attribute types underscore the need for potential targeted improvements to enhance model robustness. 

\begin{table}[t]
\centering
\caption{Accuracy (\%) across different attribute types.}
\label{tab:attribute_type}
\tiny
\begin{tabular}{@{}lcccc@{}}
\toprule
\textbf{Attribute Type} & \textbf{S-Verify} & \textbf{S-Choose} & \textbf{S-Query}\\ \midrule
SCP code & 91.1   & 85.6   & 42.9   \\
Noise & 85.8   & 83.4   & 55.9\\
Stage of infarction & 89.8   & 86.8   & 60.5\\
Extra systole & 86.4   & 75.3   & 45.2  \\ 
Heart axis & 89.3   & 89.2   & 74.2   \\
\bottomrule
\end{tabular}%
\end{table}

\begin{table}[t]
\centering
\caption{Cross-domain performance (Accuracy \%) on MIMIC-IV-ECG.}
\label{tab:cd_dataset}
\tiny
\begin{tabular}{@{}lcccc@{}}
\toprule
\textbf{Dataset} & \textbf{S-Verify} & \textbf{S-Choose} & \textbf{S-Query} & \textbf{All-S} \\ \midrule
PTB-XL & 89.0   & 84.5   & 48.6   & 41.2   \\
MIMIC w/o meta adapt & 76.3   & 49.1   & 10.4   & 13.8   \\
MIMIC w meta adapt & 89.7   & 85.7   & 39.7   & 33.8   \\ \bottomrule
\end{tabular}%
\end{table}

\subsection{Generalization on Cross-Domain Dataset}
We investigate the effect of cross-domain datasets on our model's performance under the 2-way 5-shot setting. Specifically, we evaluate the model on the MIMIC-IV-ECG dataset across different question types, with PTB-XL results provided for reference, as summarized in Table~\ref{tab:cd_dataset}. As the MIMIC-IV-ECG dataset is rather large, we randomly select 30,000 examples from its test set for evaluation to balance computational efficiency with representativeness of the dataset.

Our method demonstrates strong cross-domain capabilities, effectively working well on the MIMIC-IV-ECG dataset when meta-adaptation techniques are incorporated. With meta-adaptation, the model achieves high accuracies of 89.7\% in \textit{S-Verify} and 85.7\% in \textit{S-Choose} question types, closely aligning with the performance on the PTB-XL dataset. This highlights the effectiveness of our approach in adapting to new domains and understanding the nuances of cross-domain data.

While applying the model to the MIMIC-IV-ECG dataset without meta-adaptation results in a performance drop, the accuracy remains reasonable at 76.3\% in \textit{S-Verify} and 49.1\% in \textit{S-Choose} tasks. The incorporation of meta-adaptation significantly enhances the model's ability to generalize across domains, leading to substantial improvements in accuracy. Our method effectively leverages adaptation strategies to bridge the domain gap, enabling robust performance even when dealing with differing data distributions.

\subsection{Robustness to Question Variations}
We investigate the model's robustness to variations in question phrasing, demonstrating its ability to maintain consistent diagnostic interpretations across diversely worded queries. For example, in verification tasks (S-Verify) involving the detection of a specific SCP code, the model effectively processes semantically equivalent questions such as ``Is [SCP code] present in this ECG?'' and ``Does this ECG reveal any signs of [SCP code]?''. This indicates a capacity to generalize beyond superficial lexical variations.

Table \ref{tab:question_variation_performance} quantifies the impact of phrasing variations across different question types in a 2-way 5-shot setting. While performance modestly decreases with varied phrasing, the model retains a high degree of accuracy, demonstrating its resilience to natural language variability. This robustness is crucial for real-world applications where clinical questions are rarely phrased identically. 

\begin{table}[t]
\centering
\caption{Effect (Accuracy \%) of question expression type.}
\label{tab:question_variation_performance}
\resizebox{0.8\columnwidth}{!}{%
\begin{tabular}{@{}llll@{}}
\toprule
\textbf{Expression Type} & \textbf{S-Verify} & \textbf{S-Choose} & \textbf{S-Query} \\ \midrule
Same & 89.0   & 84.5   & 48.6   \\
Different & 86.5   & 84.7   & 42.5   \\ \bottomrule
\end{tabular}%
}
\end{table}

\subsection{Model's Capability with Reduced ECG Leads}
We investigate the influence of limiting access to ECG leads on model performance. We evaluate our approach using a reduced number of leads under a 2-way 5-shot scenario. Table~\ref{tab:masked_leads} presents the results, illustrating the effect of lead availability on accuracy across different question types.

\begin{table}[t]
\centering
\caption{Performance (Accuracy \%) with masked ECG leads.}
\label{tab:masked_leads}
\tiny
\resizebox{0.8\columnwidth}{!}{%
\begin{tabular}{@{}lccc@{}}
\toprule
\textbf{Leads} & \textbf{S-Verify} & \textbf{S-Choose} & \textbf{S-Query} \\ \midrule
I &  89.6   & 79.6   & 47.5   \\
I, II & 89.0   & 84.2   & 45.7   \\
I, II, V3 & 88.9   & 82.4   & 47.2   \\
All & 89.0   & 84.5   & 48.6 \\ \bottomrule
\end{tabular}
}
\end{table}

Using only lead I yields surprisingly high accuracy for \textit{S-Verify}, demonstrating the model's ability to effectively leverage limited information. While performance on \textit{S-Choose} and \textit{S-Query} benefits from additional leads, the strong performance with a single lead highlights the model's efficiency. Incorporating lead II further enhances performance, notably for \textit{S-Choose}, indicating the importance of this lead for choice selection tasks. While \textit{S-Query} accuracy sees a minor decrease compared to using all leads, the overall trend suggests a positive impact from incorporating more information. The inclusion of leads I, II, and V3 maintains robust performance across all question types, approaching the accuracy achieved with the full-lead scenario.

These results demonstrate that while the model benefits from access to the complete set of ECG leads, it exhibits resilience and strong performance even with limited lead availability. This adaptability suggests the model effectively learns to extract relevant features from available data, enhancing its potential for practical application in scenarios where accessing all leads might be challenging.

\begin{table}[t]
\centering
\caption{Model component ablation. Accuracy (\%) on a single-choice question type under 2-way 5-shot setting.}
\label{tab:model_comp_ablation}

\begin{minipage}[t]{0.48\columnwidth}
\centering
\scriptsize
\textbf{Multimodal fusion mapper}
\begin{tabular}{@{}lc@{}}
\toprule
\textbf{Mapper Type} & \textbf{Accuracy} \\ \midrule
Attention Based & 84.5   \\
MLP & 60.9   \\
Linear & 72.5   \\ \bottomrule
\end{tabular}
\end{minipage}
\hspace{0.2em}
\begin{minipage}[t]{0.48\columnwidth}
\centering
\scriptsize
\textbf{ECG encoder training}
\begin{tabular}{@{}lc@{}}
\toprule
\textbf{ECG Encoder} & \textbf{Accuracy} \\ \midrule
Frozen & 84.5   \\
Unfrozen & 76.7   \\ 
\bottomrule
\end{tabular}
\end{minipage}

\centering
\begin{minipage}[t]{0.8\columnwidth}
\centering
\scriptsize
\textbf{Meta-knowledge impact}
\begin{tabular}{@{}lc@{}}
\toprule
\textbf{Variants} & \textbf{Accuracy} \\ \midrule
w meta knowledge & 84.5   \\
w/o meta knowledge & 0.3   \\ 
\bottomrule
\end{tabular}
\end{minipage}
\end{table}

\subsection{Architectural Components Ablation}
\paragraph{Multimodal Fusion Mapper.}
We investigated the efficacy of three distinct multimodal fusion mappers: attention-based, linear, and multilayer perceptron (MLP) (See Section~\ref{sec:multimodal_fusion_mapper}). In Table~\ref{tab:model_comp_ablation}, we see that the attention-based mapper consistently demonstrated superior performance, achieving an accuracy of 84.5\%, compared to 60.9\% for the MLP mapper and 72.5\% for the linear mapper. This suggests that the attention mechanism's ability to dynamically weigh and integrate modality-specific information is crucial for effective multimodal reasoning in this context. Consequently, we employed the attention-based mapper as the foundation for subsequent ablation experiments.

\paragraph{Freezing ECG Encoder Parameters.}
We investigate the effects of freezing the pre-trained ECG encoder parameters on few-shot learning performance in Table~\ref{tab:model_comp_ablation}. Specifically, we compare the performance of a model with a frozen ECG encoder against a model where the encoder parameters are allowed to be fine-tuned during training. This evaluation uses the single-choice question type in a 2-way 5-shot setting. Freezing the ECG encoder parameters yields a higher accuracy of 84.5\%, compared to 76.7\% for the unfrozen encoder. This result suggests that for few-shot learning in this context, leveraging the pre-trained representations without further fine-tuning is more effective. Furthermore, freezing the encoder parameters reduces computational overhead and mitigates the risk of overfitting on the limited few-shot data.

\paragraph{Meta-Knowledge Incorporation.}
Incorporating meta-knowledge significantly improves performance on few-shot learning tasks. Meta-knowledge refers to information about the learning process itself, such as patterns or strategies learned from previous tasks that can be applied to new tasks with limited data \citep{finn2017model}. Table~\ref{tab:model_comp_ablation} provides these results, where our model achieved 84.5\% accuracy on single-choice questions when leveraging meta-knowledge. Accuracy dropped drastically to 0.3\% without learning meta-knowledge, highlighting the critical role of prior information for improved understanding and decision-making in few-shot scenarios.\citep{rafiei2024meta}

\paragraph{Impact of Prompt Format on Model Performance.}
We investigate the influence of prompt variations on model performance for few-shot ECG-language question answering. Specifically, we evaluate three prompt variants (P-A, P-B, and P-C) using a 2-way 5-shot learning paradigm on single-choice questions. Table~\ref{tab:prompts} summarizes the results and demonstrates a clear impact of prompt structure on accuracy. The most effective prompt, P-A (``question: " + question + ``answer: "), achieves the highest accuracy (84.5\%). This structured format provides explicit cues for the question and expected answer, facilitating the model's comprehension and response generation. In contrast, the simpler P-B variant (question only) results in a lower accuracy of 77.4\%, suggesting the importance of contextual cues present in P-A. The P-C variant (question + "the answer can be both, none or in question ") achieves an intermediate accuracy of 80.1\%. While the added clarification in P-C might be beneficial in certain scenarios, it does not improve performance compared to the structured approach of P-A. Our findings underscore the critical role of prompt format in optimizing large language model performance for few-shot question answering tasks.

\begin{table}[t]
\caption{Effect (Accuracy \%) of varying prompt structures.}
\label{tab:prompts}
\resizebox{\columnwidth}{!}{%
\begin{tabular}{@{}lc@{}}
\toprule
\textbf{\large{Prompt Variants}} & \textbf{\large{Accuracy}}\\ \midrule
P-A (``question: '' + question + ``answer: '') & 84.5   \\
P-B (question) & 77.4   \\
P-C (question + ``the answer can be both, none or in question'') & 80.1   \\ \bottomrule
\end{tabular}%
}
\end{table}
\section{Conclusion}
In this work, we introduce a LLM-agnostic multimodal meta-learning framework for few-shot ECG-language question answering, addressing the critical challenges of limited labeled data and evolving task distribution in ECG interpretation. Our framework seamlessly integrates ECG signals with text queries through a trainable multimodal fusion mapper. The empirical evaluation demonstrates superior generalization performance across a range of language models, diverse few-shot learning scenarios, and varying question types. These results underscore the potential of our framework to enhance clinical practice by enabling rapid adaptation to new tasks and patient populations.
Our method can be easily extended to multiple ECG comparisons by incorporating multiple ECG prefixes in the LLM decoder. Future research could explore incorporating vision modality (e.g., chest X-ray images) to develop more comprehensive models.
Additionally, investigating different ECG encoder variants to enhance model robustness across different patient demographics, hospitals, and ECG devices. Leveraging larger language models (LLMs), and integrating more established few-shot learning methods over multiple, randomly seeds could further improve performance and generalizability.

\section*{Acknowledgments}
We acknowledge the use of the Dutch National Supercomputer Snellius for essential computational tasks.

\bibliography{main}

\clearpage
\onecolumn
\appendix
\label{apd:first}

\section{Implementation Details}

\subsection{ECG Encoder Pretraining Parameters}
\label{apd:Pretraining}
During pretraining, we apply random lead masking by independently masking each lead with a probability of $p = 0.5$, enhancing the model's robustness to missing or corrupted leads. The ECG encoder is trained using the Adam optimizer with a learning rate of $5 \times 10^{-5}$ for 200 epochs. 

\subsection{Dataset Pre-Processing Details} 
\label{apd:Dataset}
Meta-training dataset \( D_{\text{meta-train}} \) and meta-testing dataset \( D_{\text{meta-test}} \) are composed of data points \( (x_i, q_i, a_i) \) drawn from their respective sets of classes \( C_{\text{meta-train}} \) and \( C_{\text{meta-test}} \), where \( C_{\text{meta-train}} \cap C_{\text{meta-test}} = \emptyset \), ensuring disjoint class sets for training and testing. For each question type, the data were split into 80\% for training and 20\% for testing. 

\subsection{Multimodal Fusion Module Architecture Parameters}
\label{apd:multimodal_fusion_mapper}

Attention-based Mapper utilizes the multi-head attention mechanism with 8 heads, 4 layers, and a dropout rate of 0.5. Similarly, the Linear Mapper applies a linear transformation, i.e., a single-layer model. Furthermore, the MLP Mapper utilizes a feed-forward neural network with $3$ layers and ReLU activation with a dropout rate of 0.5 to prevent overfitting.

\subsection{Training \& Inference Procedures Parameters}
\label{apd:Training infer}

The meta-level outer learning rate is set to \(5 \times 10^{-4}\), while the task-level inner update learning rate is 0.05. The inner update step in meta-learning refers to the process of adapting the model’s parameters to a specific task during inner iteration based on the support set \citep{finn2017model}. The task-level inner update steps are set to the default value of 5, and the update steps for fine-tuning are also set to the default value of 15. Due to resource limitations, we train models for one epoch (which roughly takes over a duration of 1-2 days), utilizing a step size of 10,000 split across NVIDIA H100 GPUs. We keep both the ECG encoder and language model frozen, unless mentioned otherwise. Implementation details like seeds will be released with our code.

\section{Additional Figures and Analysis}
This appendix contains additional figures and analysis that provide further insights into our experiments and results. The following subsections detail class formation, attribute distribution, meta-learning processes, and qualitative analysis of ECG-related question answers.

\subsection{Meta-Training and Meta-Testing Processes}
\label{apd:second}
\begin{figure*}[htbp]
\centering
\includegraphics[width=\textwidth]{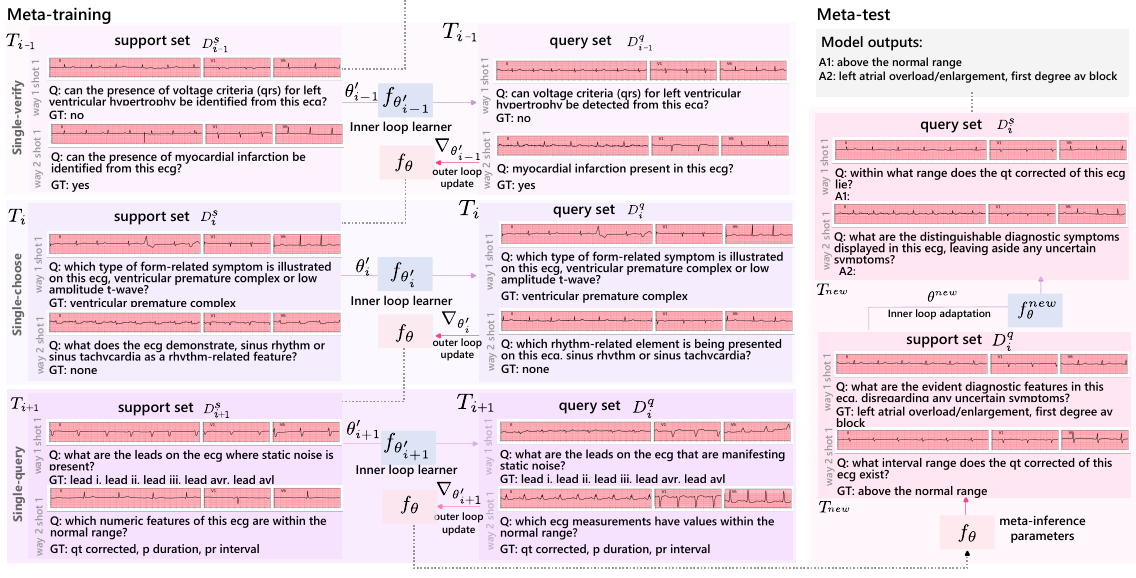}
\caption{Illustration of the meta-training and meta-testing processes of our approach.}
\label{fig:meta_training_overview}
\end{figure*}

Following the Model-Agnostic Meta-Learning (MAML)~\citep{finn2017model} structure, we train the model on a variety of ECG question-answering tasks in the meta-training phase to make it optimize the model’s ability to quickly adapt to new tasks with minimal data. We highlight how the model's parameters are adjusted across multiple training episodes, leading to improved accuracy in the few-shot settings presented in the study. We demonstrate the key components of the process that are critical for understanding how the models adapt in figure \ref{fig:meta_training_overview}.

\subsection{ECG-Related Question Answering: Qualitative Analysis}
\label{apd:third}
\begin{figure*}[t]
\centering
\includegraphics[width=\textwidth]{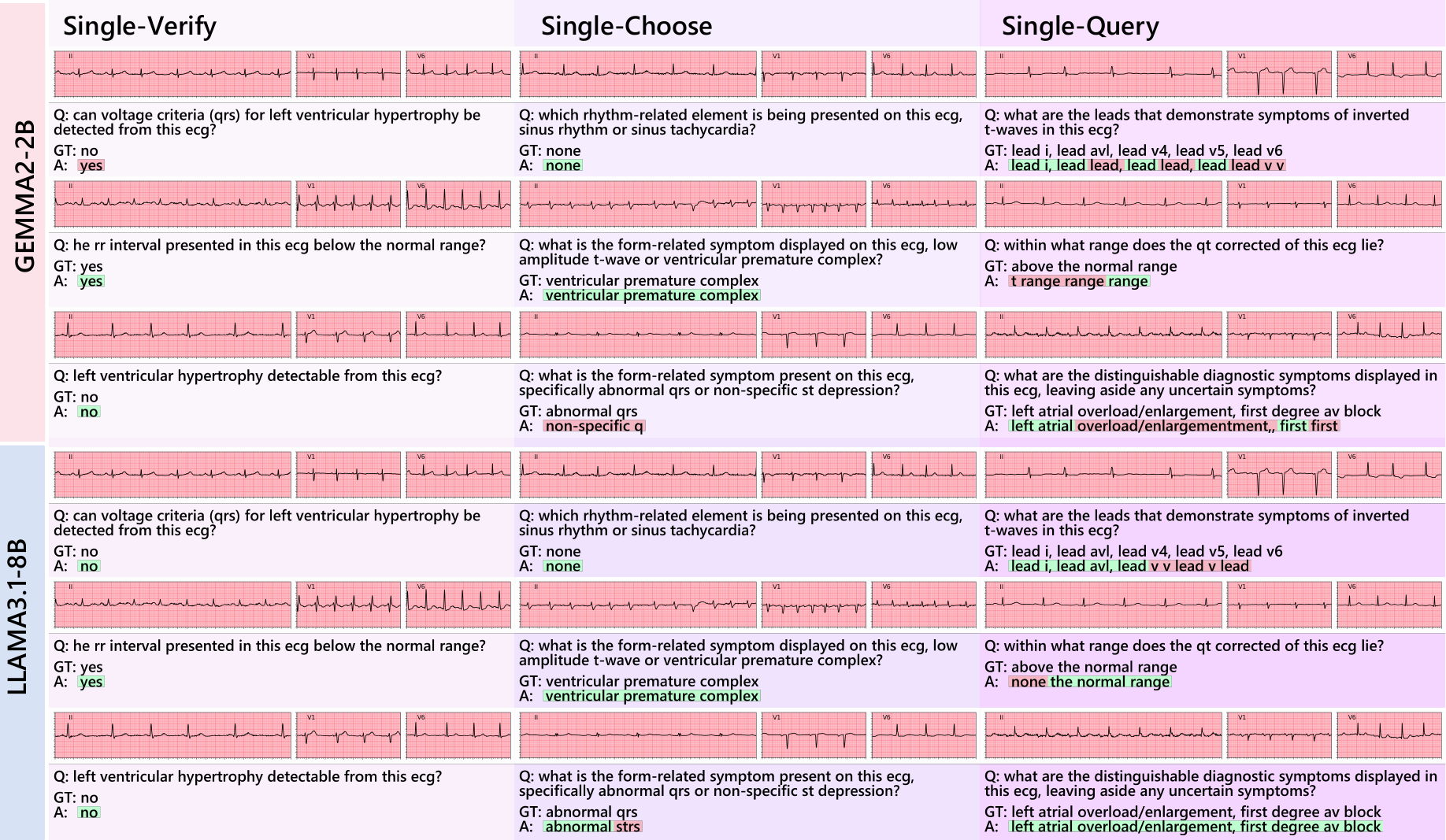}
\caption{Qualitative analysis of Gemma-2-2B and Llama-3.1-8B models across randomly selected questions.} 
\label{fig:ecg_qa_examples}
\end{figure*}

This figure \ref{fig:ecg_qa_examples} presents qualitative results comparing two models, Gemma-2-2B and Llama-3.1-8B, on single-verify, single-choose, single-query, 3 ECG-related question types. The analysis helps in understanding the models' performance and their ability to handle various question forms.

\end{document}